\documentclass[letterpaper, 10 pt, conference]{ieeeconf} 
\IEEEoverridecommandlockouts                              
\usepackage{graphics} 
\usepackage{epsfig} 
\usepackage{mathptmx} 
\usepackage{times} 
\usepackage{amsmath} 
\usepackage{amssymb}  
\usepackage{url}
\usepackage{soul}
\usepackage{tikz} 
\let\labelindent\relax
\usepackage{enumitem}
\usepackage{hyperref}
\usepackage{subfigure}
\usepackage[normalem]{ulem}

\hypersetup{colorlinks=true}

\begin{document}

\title{\LARGE \bf
    Autonomy and Perception for Space Mining
}

\author{
    Ragav Sachdeva$^{\dagger,*}$,
    Ravi Hammond$^{\dagger,*}$,
    James Bockman$^{\dagger}$,
    Alec Arthur$^{\dagger}$,
    Brandon Smart$^{\dagger}$,
    Dustin Craggs$^{\dagger}$, ~\\
    Anh-Dzung Doan$^{\dagger}$,
    Thomas Rowntree$^{\dagger}$,
    Elijah Schutz$^{\dagger}$,
    Adrian Orenstein$^{\dagger}$,
    Andy Yu$^{\dagger}$,
    Tat-Jun Chin$^{\dagger,\ddagger}$,
    Ian Reid$^{\dagger}$ ~\\
    \thanks{$^{*}$These authors contributed equally to this work.}
    \thanks{$^{\dagger}$Australian Institute for Machine Learning and Andy Thomas Centre for Space Resources, The University of Adelaide.}
    \thanks{$^{\ddagger}$SmartSat CRC Professorial Chair of Sentient Satellites.}
}

\maketitle
\thispagestyle{empty}
\pagestyle{empty}

\begin{abstract}
Future Moon bases will likely be constructed using resources mined from the surface of the Moon. 
The difficulty of maintaining a human workforce on the Moon and communications lag with Earth means that mining will need to be conducted using collaborative robots with a high degree of autonomy. 
In this paper, we describe our solution for Phase 2 of the NASA Space Robotics Challenge, which provided a simulated lunar environment in which teams were tasked to develop software systems to achieve autonomous collaborative robots for mining on the Moon. 
Our \textit{3rd place and innovation award winning solution} shows how machine learning-enabled vision could alleviate major challenges posed by the lunar environment towards autonomous space mining, chiefly the lack of satellite positioning systems,  hazardous terrain, and delicate robot interactions. 
A robust multi-robot coordinator was also developed to achieve long-term operation and effective collaboration between robots\footnote{A recording of our robots in action is available at~\cite{yt_overall}.}.
\end{abstract}

\section{Introduction}

The need to transport resources from Earth is a serious obstacle to space exploration that must be addressed as a precursor to sustainable deep space missions. 
In-Situ Resource Utilisation (ISRU), where resources are extracted on other astronomical objects and exploited to support longer and deeper space missions, has been proposed as a way to mitigate the need to carry resources from Earth~\cite{isru_utilisation_for_mnm_exploration}. 

The difficulties of building a large-scale human presence in space and the lack of real-time interplanetary communication means that mining on planetary bodies (primarily the Moon and Mars) will have to depend on robots with a high level of autonomy~\cite{thangavelautham2020autonomous, neale2011space}. Although there exist semi-automated systems for mining on Earth~\cite{ghodrati2015reliability}, they are supported by mature infrastructure such as global navigation satellite systems (GNSS), well-maintained roads, ready access to fuel, and maintenance.
These facilities will not be available at the onset of space mining missions, where robots will need to contend with hazardous terrain, lack of accurate positioning systems, limited power supply, and many other difficulties~\cite{mikrin2019satellite, schwendner2014space, sasiadek2014space, cheung2017situ, wong2017adaptive}.
Indeed, space robotics has been identified by NASA as a \emph{Centennial Challenge}.

For risk and economic reasons, space mining will likely utilise a fleet of heterogeneous robots that must collaborate to accomplish the goal. 
This accentuates the difficulties alluded to above; apart from being able to navigate in an unstructured environment and avoid obstacles without accurate satellite positioning, a robot must also manoeuvre and interact with other robots without causing damage. 
This argues for a high degree of intelligence on each agent and a robust multi-robot coordination system to ensure long-term operation.

\begin{figure}[t]\centering
    \includegraphics[width=0.9\linewidth]{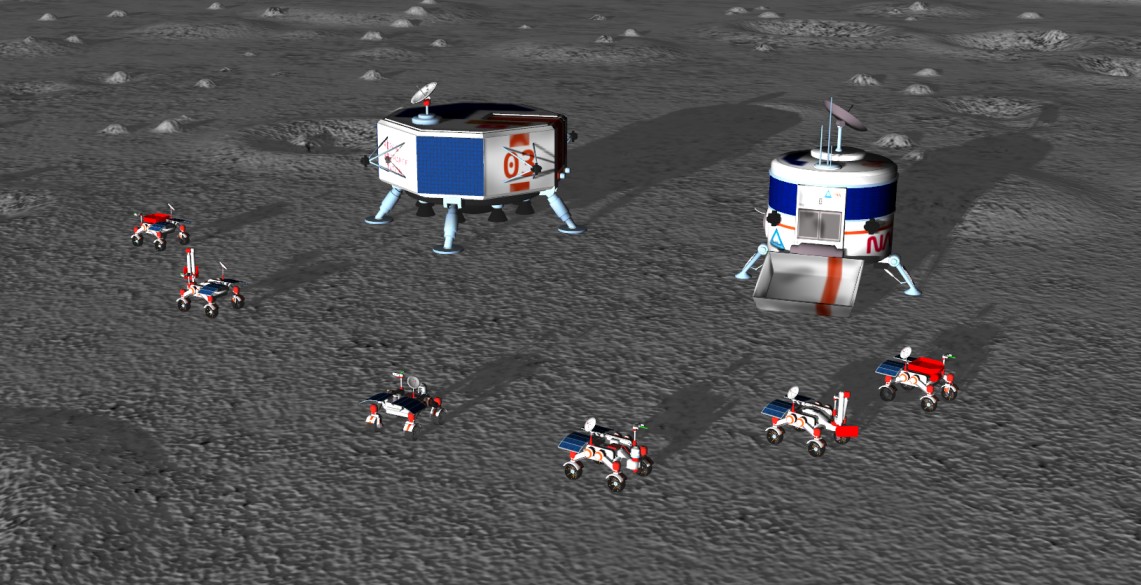}
    \caption{Gazebo simulated lunar environment with rovers in SRCP2.}
    \label{fig:lunar_env}
\end{figure}

In this paper, we describe a system to address some of the key challenges towards autonomous robots for collaborative space mining: lack of satellite positioning systems, navigation in hazardous terrain, and the need for delicate robot interactions.
Specifically, we describe the main components of our solution for the \textit{NASA Space Robotics Challenge Phase 2 (SRCP2)}~\cite{challenge2020srcp2}, wherein a simulated lunar environment that contained a heterogeneous fleet of rovers was provided; see Fig.~\ref{fig:lunar_env}. 
The goal was to develop software to enable the rovers to autonomously and collaboratively find and extract resources on the Moon. 
Our \textit{3rd place and innovation award winning solution}, extensively employed machine-learning based robotic perception to accomplish accurate localisation, semantic mapping of the lunar terrain, and object detection to facilitate accurate close range manoeuvring between rovers.

In the rest of the paper, we further introduce SRCP2, briefly describe our overall solution, and detail our robotic vision-enabled algorithms and their results on the problems above. 

\section{NASA Space Robotics Challenge}\label{sec:srcp2}

In SRCP2~\cite{challenge2020srcp2}, a Gazebo simulated lunar environment that contained several rovers and two lunar landers (``base stations'') was provided; see Fig.~\ref{fig:lunar_env}. 
Competitors were tasked with developing software that enables the rovers to autonomously find, excavate, and retrieve resources (volatiles) in the lunar regolith. 

\begin{figure}[th]\centering
    \includegraphics[width=0.9\linewidth]{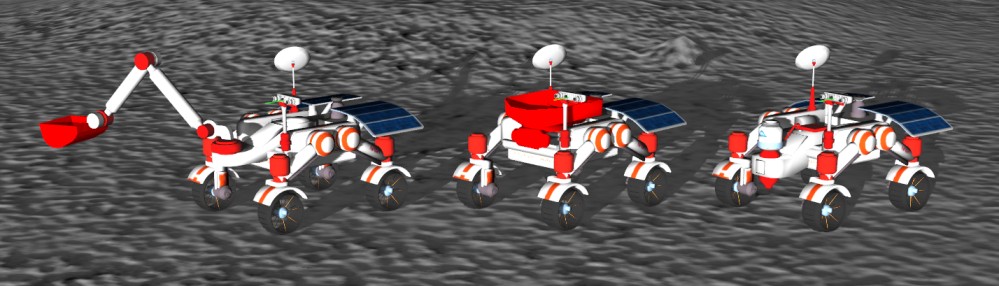}
    \caption{Rover classes in SRCP2 (excavator, hauler, and scout).}
    \label{fig:rovers}
\end{figure}

The main features/constraints of the challenge are:
\begin{itemize}[leftmargin=1em]
    \item Resources are scattered across the 200m$\times$200m map with no prior information about their locations. Hence, the resources must be found by exploring the environment.
    \item Hazardous terrain comprising mounds, craters and hills, which can cause the robot to slip, disorient, or flip. Therefore, obstacles must be avoided during navigation.
    \item Absence of a global positioning system. Each rover is allowed to query its global position only once from the simulator (e.g., for initialisation); thus, the rovers need to self-localise. In addition, the positions of the base stations are neither supplied nor retrievable from the simulator.
    \item The base stations comprise a processing plant, where all the excavated resources must be deposited, and a recharge station, to rapidly restore rover batteries.
    \item There are three types of rovers---scout, excavator and hauler (see Fig.~\ref{fig:rovers})---that have complementary specialisations. The scout has a volatile sensor to locate resources and is the fastest rover, the excavator has an arm for digging, and the hauler has a bin to haul the resources back to the processing plant. In addition, each rover is equipped with an IMU, stereo cameras, and a 2D LiDAR.
    \item The challenge allows fleets to comprise of any combination of the rover types up to a maximum of six units.
    \item The final score is the number of volatiles deposited in the processing plant during a 2-hour simulation run.
\end{itemize}


\section{Overview of our solution}\label{sec:overview}

Our paper focuses on the role of robotic perception and multi-robot coordination towards localisation, navigation, robot interaction and coordination. It is nevertheless useful to first provide an overview of our solution to help conceptually connect the major components to be described later.

Our solution utilises two scouts, two excavators, and two haulers separated into two largely independent teams. Each team consists of one instance of each rover type. At initialisation the poses of all rovers and base stations are established on a common world coordinate frame (Sec.~\ref{sec:localisation}). Upon successful initialisation, the on-board localisation algorithm of each rover is invoked.

The scouts then follow a spiral semi-circle search pattern centred at the base stations to discover volatiles, which prioritises the discovery of deposits closer to base. 
Meanwhile, the excavator and the hauler of each team follow their respective scout, ready to extract volatiles as soon as they are found. 
Throughout the journey, the rovers continuously generate semantic understanding of their surroundings through the camera to conduct real-time obstacle avoidance (Sec.~\ref{sec:navigation}).

During exploration, the scout continuously monitors its volatile sensor that returns a noisy measurement of distance to the centre of a volatile within a 2m radius. 
When a deposit is detected, the scout attempts to pinpoint the location of the deposit. It does this by first rotating on the spot to align its orientation with the direction of the detected volatile (via gradient descent in conjunction with a Savitzky–Golay filter~\cite{savgol}), and then repeating the process while driving forward. 
Upon successful volatile detection, the scout waits for the excavator and hauler to rendezvous with it (Sec.~\ref{sec:interactions}). 
Once a safe park has been is reached, the scout continues exploring while the excavator and hauler begin mining.

The excavator repeatedly digs for volatiles and dumps them into the hauler's bin using object detection on the camera feed to locate the hauler (Sec.~\ref{sec:scene_understanding}), and LiDAR to accurately infer distance from the hauler's bin to the excavator chassis (Sec.~\ref{sec:interactions}). The excavator then returns to following the scout, and the hauler may return to the processing plant if its bin is full. The above is repeated until both teams exhaust all resources in their respective domain.

When a rover's battery level is low, it pauses its current task and returns to the repair station. While approaching the base stations, accumulated error in on-board localisation is zeroed by estimating the rover pose with respect to the base stations (Sec.~\ref{sec:localisation}) when the latter are in view.

Our solution was able to consistently and continuously operate in 2-hour simulation runs of SRCP2; see~\cite{yt_overall} for a video recording. 
In the following, we will explain in more detail how we accomplished localisation, navigation, robot interaction, and coordination, particularly the robotic vision algorithms that underpin the former three components.

\section{Localisation}\label{sec:localisation}

\begin{figure*}[ht]\centering
    \includegraphics[width=0.8\linewidth]{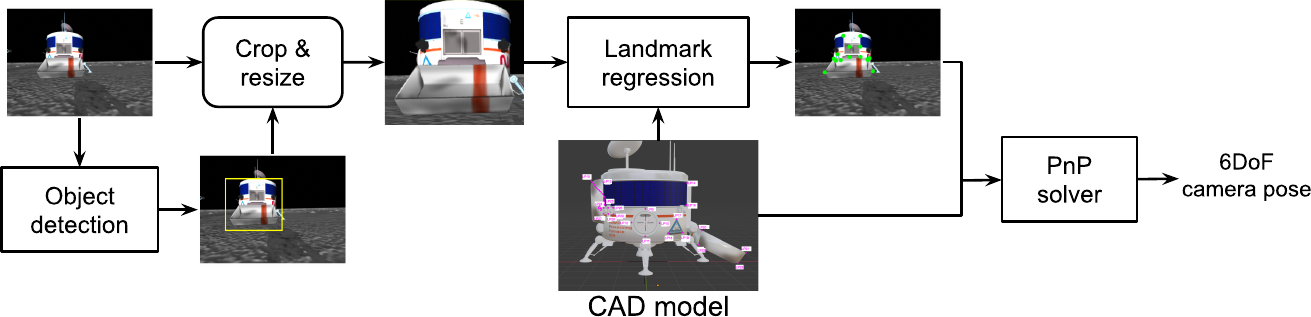}
    \caption{Pose estimation pipeline to calculate the relative pose between base stations and rover.}
    \label{fig:pose_pipeline}
\end{figure*}

Accurate localisation---estimating position and orientation within the operating environment---is fundamentally important to autonomous robots~\cite{thrun2002probabilistic}. 
Localisation techniques can be broadly classified into active and passive methods.
Active methods generally involve direct communication of signals that facilitate localisation. Examples include RF beacons, WiFi positioning, RFID positioning and GNSS.

Passive methods utilise onboard sensors to generate relative measurements between the robot and the environment to estimate position. A basic technique is to conduct dead reckoning using interoceptive sensors such as wheel encoders and an IMU to incrementally track the robot motion using Bayesian filtering. 
However, dead reckoning is subject to drift, hence the filter must be periodically reset using extra information such as celestial positioning~\cite{yang2014simultaneous, zhan2021adaptive}, fiducial markers~\cite{bou2019geometry, zenatti2016optimal}, or image matching~\cite{li2007rock, hu2018cvm}.

Simultaneous localisation And Mapping (SLAM) is regarded as a state-of-the-art (SOTA) passive localisation approach. 
In addition to tracking robot motion, SLAM techniques incrementally build a map of the environment using the sensor percepts. 
This allows the robot to relocalise itself in the environment (so called ``loop closing'') and remove drift by redistributing accumulated error through all variables in the system. 
A notable instance of SLAM is visual SLAM (VSLAM), whereby the primary sensor is a camera~\cite{davison2007monoslam}. 
SOTA VSLAM algorithms~\cite{mur2017orb, sumikura2019openvslam} detect and map visually salient features or keypoints in the environment.

A practical robot localisation scheme will likely use a combination of active and passive methods. 
It is worthwhile to point out that existing Earth-centric GNSS will unlikely be sufficient for accurate localisation on the Moon~\cite{manzano2014use, mikrin2019satellite}.

\subsection{Our localisation technique for SRCP2}

\begin{figure}[ht]\centering
    \subfigure{\includegraphics[width=0.7\columnwidth]{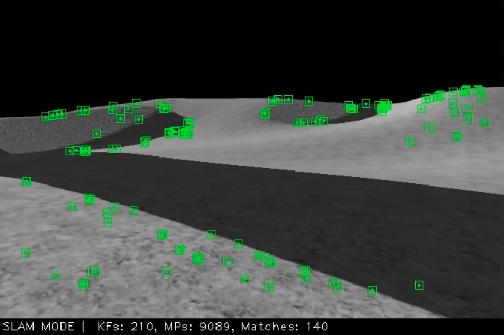}}
    \caption{ORB keypoints detected in the first person view (FPV) of a rover. Notice the relatively texture-poor terrain and strong contrast between bright and shadowed regions. Also, few keypoints are detected in the shadow.}
    \label{fig:orbslam}
\end{figure}

Active localisation functionalities and positioning markers are not provided in SRCP2. 
Realistic star fields are also not rendered in the simulator. 
Our initial investigation also showed that VSLAM~\cite{mur2017orb} is brittle~\cite{wang2017survey} in the simulated lunar environment, possibly due to the feature-poor textures used to render the map, significant brightness contrast and strong shadowing, which reduce the ability to repeatably detect and match keypoints; see Fig.~\ref{fig:orbslam}. Investigations also showed that LiDAR only produced 5 measurements, and readings were too noisy to be used for localisation.

\begin{figure}[ht]\centering
    \subfigure{\includegraphics[width=0.44\linewidth]{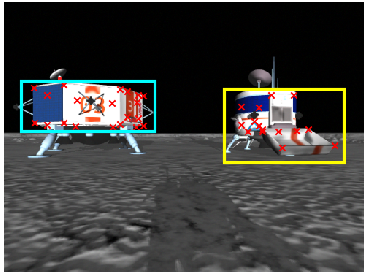}}
    \subfigure{\includegraphics[width=0.54\linewidth]{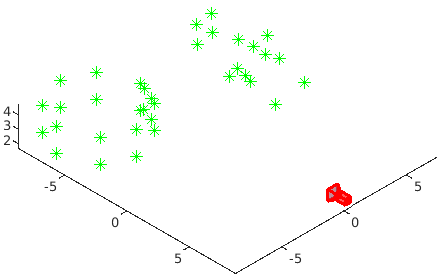}}
    \caption{Relative pose estimation between rover and observed base stations. See~\cite{yt_pose} for a video recording of our pose estimation results.}
    \label{fig:pose_results}
\end{figure}

Given the above findings, we developed a localisation solution that relies on extended Kalman filtering (EKF)~\cite{moore2016generalized,  kilic2020team} of velocity estimates from the wheel odometry and onboard IMU of a rover. 
Each rover independently tracks its position. 
To reset drift, we perform visual pose estimation of the base stations.
Specifically, as a base station appears in the FOV of the rover camera (per normal return to base runs; see Sec.~\ref{sec:overview}), the 6DoF relative pose between the base station and rover is estimated; see Fig.~\ref{fig:pose_results}. 
Given the absolute pose of the base station (initialised according to Sec.~\ref{sec:init}), the relative absolute pose of the rover is inferred and used to reset the EKF. 
We describe the pose estimation pipeline and initialisation next.

\subsection{Visual pose estimation}

To relocalise rovers we adopted the satellite pose estimation pipeline of Chen et al.~\cite{chen2019satellite}; see Fig.~\ref{fig:pose_pipeline}. 
A deep neural network (DNN) (specifically a YOLOv5~\cite{glenn_jocher_2021_4679653} object detector) detects the bounding box of base stations in the input image. 
Given a crop of a base station, a second DNN (a combination of HRNet~\cite{sun19deep} and DSNT~\cite{nibali2018numerical}) predicts the coordinates of predetermined landmarks in the image. 
This gives rise to 2D-3D correspondences between the image and the 3D model of the base station, which is fed to a robust perspective-n-point (PnP) solver~\cite{fischler81} to compute the relative pose.
The pipeline inference was performed at 10 FPS on RTX 2080 (constrained by the simulation).
To train the object detector, we collected 30,000 images from the FPV of rovers and labelled them with ground truth bounding boxes containing base stations.  To train the landmark predictor, we manually chose, based on visual saliency, 35 points on the CAD model of the base stations provided by NASA, and labelled the pixel locations of the points in the training images. DNN training was done using PyTorch and the YOLOv5 framework~\cite{glenn_jocher_2021_4679653}.

\begin{figure*}[ht]
    \centering
    \includegraphics[width=0.8\linewidth]{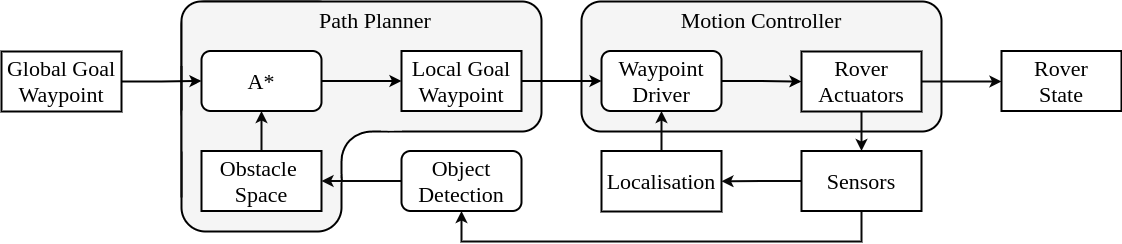}
    \caption{Navigation framework that combines the path planner and waypoint driver. In order to reach the global goal waypoint while avoiding obstacles, the path planner creates a series of local goals. The motion controller drives the rover's actuators to steer it to the next local goal.}
    \label{fig:navigation_framework}
\end{figure*}

\subsection{Initialisation}\label{sec:init}

As mentioned in Sec.~\ref{sec:srcp2}, while the ground truth absolute pose of rovers can be accessed once via an API call to the simulator, no such facility exists for the base stations. 
To localise the base stations, we developed a routine whereby, upon spawning of the environment and objects, a scout rotates about the azimuth (achievable due to the differential drive) to attempt to find the base stations (note that all rovers are spawned close to the base stations; see Fig.~\ref{fig:lunar_env}). 
As soon as a base station is in view, its relative pose is estimated. The ground truth absolute pose of the scout is queried and propagated using this relative pose to estimate the absolute pose of the base station.


\section{Navigation}\label{sec:navigation}

The uneven lunar terrain is hazardous for rovers due to the presence of mounds, craters, and hills. 
To accomplish autonomous space mining, rovers need to be able to avoid obstacles while automatically navigating the environment.

There has been recent interest in solving the navigation problem using end-to-end deep learning approaches~\cite{a3cnav, seymour2021maast, chaplot2020learning}. 
However, these methods typically train their models on complex, feature-rich environments and either use simplistic motion models or assume that the agents are equipped with accurate satellite positioning. 
In addition, they suffer from intrinsic problems of learning methods including high demand of training data~\cite{chaplot2020learning}, overfitting of the environments, and lack of explainability~\cite{xiao2020motion}.

\begin{figure}[ht]\centering
    \includegraphics[width=0.9\linewidth]{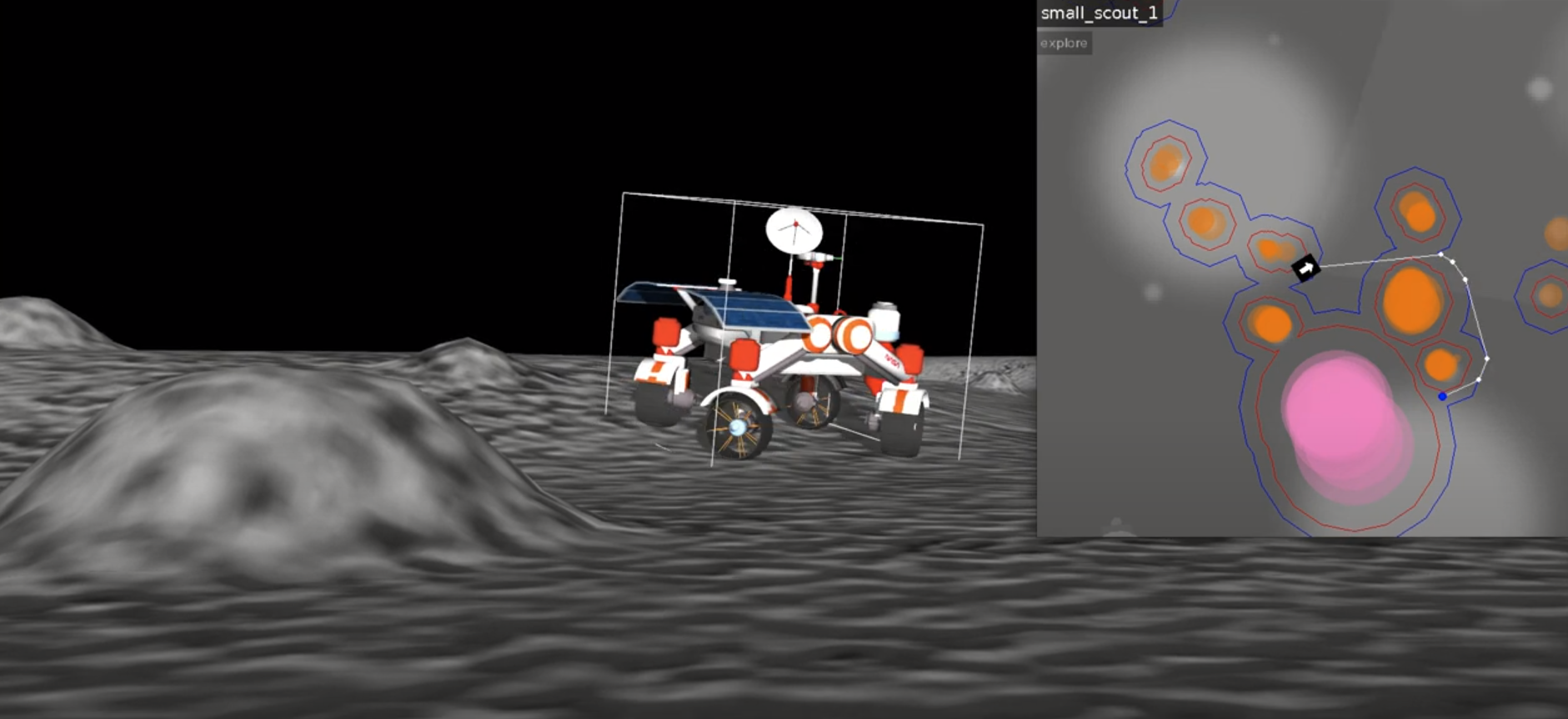}
    \caption{Real-time semantic scene understanding for obstacle avoidance in navigation in our SRCP2 solution. In the map inset, pink indicates craters while orange indicates mounds. See the full video at~\cite{yt_semantic}.}
    \label{fig:small_scout_1_minimap_detections}
\end{figure}

Keeping reliability and robustness in mind, we developed a navigation approach for SRCP2 based on classical methods~\cite{xiao2020motion,lavalle2006planning} that is informed by robotic vision. 
Similar to classical navigation, we use a hierarchical approach consisting of a path planner and a motion controller that work in tandem; see Fig.~\ref{fig:navigation_framework}. 
To achieve real-time obstacle avoidance, a key component of our navigation framework is to perform semantic understanding of the local environment of the rover. 
The map inset in Fig.~\ref{fig:small_scout_1_minimap_detections} illustrates our semantic scene understanding, while~\cite{yt_semantic} is a video recording that highlights our navigation system for SRCP2. 
We provide more details of our navigation approach in the following.

\subsection{Semantic scene understanding}\label{sec:scene_understanding}

\begin{figure}[ht]\centering
    \includegraphics[width=0.8\linewidth]{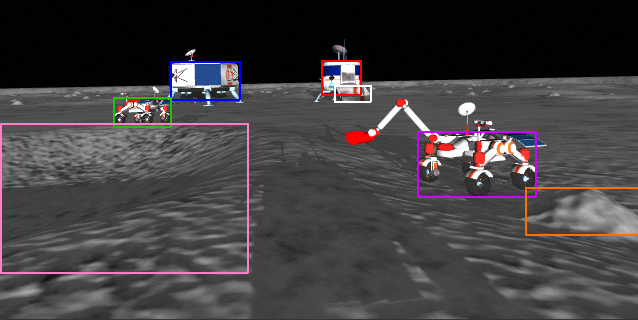}
    \caption{Detection of several object classes (processing plant, repair station, hauler, excavator, crater, mound) in the FOV of a scout.}
    \label{fig:YOLOv5_example_FPV_detecting_objects}
\end{figure}

We used object detection and depth estimation to separately generate semantic local scene understanding for each rover, i.e., the identity and positions of select objects close to the rover. 
A YOLOv5~\cite{glenn_jocher_2021_4679653} object detector was trained to detect base stations (processing plant and repair station), other rovers (scouts, excavators, haulers), mounds, and craters in the FOV of the rover camera; see Fig.~\ref{fig:YOLOv5_example_FPV_detecting_objects}.
As the development and final grading environments differed by only seed randomisation, the model didn't need to generalise any further.
For each object detected by a rover, its distance to the rover is determined using stereo-depth estimation~\cite{faugeras1993real}, resulting in a real-time local semantic map (unique to the rover) as shown in Fig.~\ref{fig:small_scout_1_minimap_detections}. 
To train the object detector, approximately 10,000 images labelled with ground truth bounding boxes were used. 
Training and implementation was done using PyTorch and the YOLOv5 training pipeline.

As EKF accumulates error, the relative position of the previously detected objects become inaccurate. 
To address this, we introduced a time-to-live (TTL) value for each object, dictating how long it should persist in the map before being removed. 
To avoid situations where rovers are stationary for long periods of time and all the objects in their periphery expire, we pause the TTL countdown for objects if the rover is not moving. Additionally, we maintain a 7m radius about the rover in which any objects will have their TTL countdown indefinitely paused, which was particularly useful when maneuvering around obstacles outside the rover's FOV.

\subsection{Path planning and motion control}

Here we briefly outline our navigation framework that relies extensively on the ability to generate real-time semantic understanding of the local surrounding of each rover. 
Akin to the classical methods~\cite{xiao2020motion,lavalle2006planning}, our system has two levels of planning---path planner and motion controller; see Fig.~\ref{fig:navigation_framework}. 
The path planner, given a final destination and the local (obstacle) map, generates a series of unobstructed waypoints that the rover can follow to reach the desired destination. 
This path is computed by using the A* shortest-path algorithm~\cite{hart1968formal} on a fully-connected graph composed of points on the boundary of the obstacles in the local map (excluding edges that intersect with obstacles).
The motion controller, given an unobstructed waypoint, is then responsible for generating control signals to efficiently move the rover from its current position to the goal waypoint.


\section{Robot interactions}\label{sec:interactions}

Collaboration between heterogeneous robots is essential in SRCP2. Here, we describe two main aspects of robot interactions (rover rendevzous and excavation/dumping) that employ robotic vision extensively in our solution.

\subsection{Rover rendezvous}\label{sec:rendezvous}

\begin{figure}[ht]\centering
    \subfigure{\includegraphics[height=0.38\columnwidth]{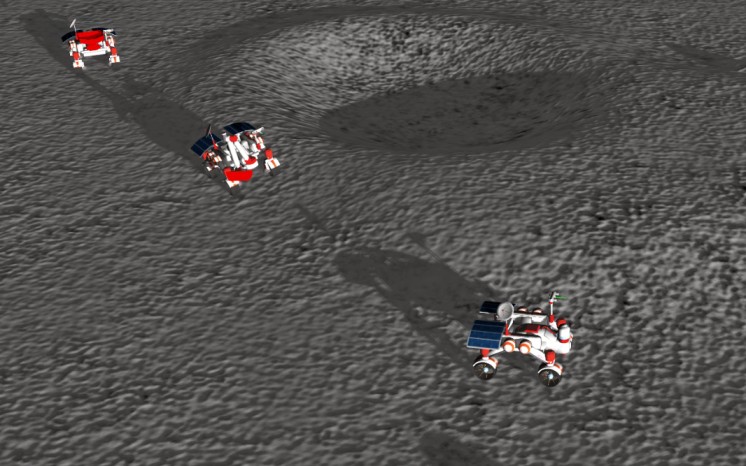}}
    \subfigure{\includegraphics[height=0.38\columnwidth]{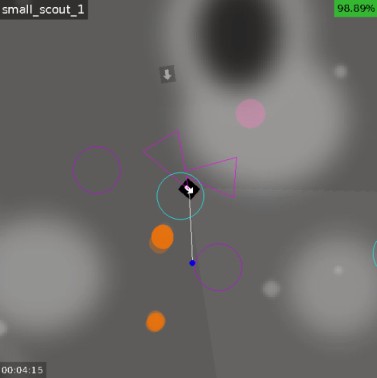}}
    \caption{Three rovers (scout, excavator and hauler) at the onset of rendezvous. The targeted configuration is shown by the purple triangle pair in the local semantic map of the scout. See~\cite{yt_rendezvous} for a video of the process.}
    \label{fig:rendezvous}
\end{figure}

Rover rendezvous is the activity whereby a scout, an excavator and a hauler come into close proximity to extract and deposit volatile in the regolith.
Rendezvous is extremely delicate since any error in the process will cause collision, resulting in an increased EKF drift or damage. Fig.~\ref{fig:rendezvous} depicts the onset of rover rendezvous in our solution.

A major obstacle to rendezvous is potential localisation inaccuracies of the rovers (by up to 5m), which we mitigate using visual guidance in the close range. 
First, once a scout finds a volatile, it pauses on the spot to function as a ``marker'' for the resource. 
The scout then broadcasts the (estimated) location to the rovers along with an obstacle-free parking configuration, defined as a triangle pair; see Fig.~\ref{fig:rendezvous}. 
The excavator then approaches the scout based on the broadcasted position estimate. 
When the excavator is within 10m of the scout, it stops using the broadcasted scout position, and engages its camera and objector detector (see Fig.~\ref{fig:YOLOv5_example_FPV_detecting_objects}) to visually locate the scout. 
The predicted bounding box of the target rover in conjunction with stereo-depth are used to estimate the precise location of the volatile relative to the excavator. 
The scout then safely departs the dig-site, allowing the excavator to park itself in front of the deposit, as per the safe ``triangle", ready for extraction.
Subsequently and in a similar fashion, the hauler approaches the excavator to complete the rendezvous.

\subsection{Excavation and dumping}

\begin{figure}[ht]\centering
    \subfigure{\includegraphics[height=0.35\columnwidth]{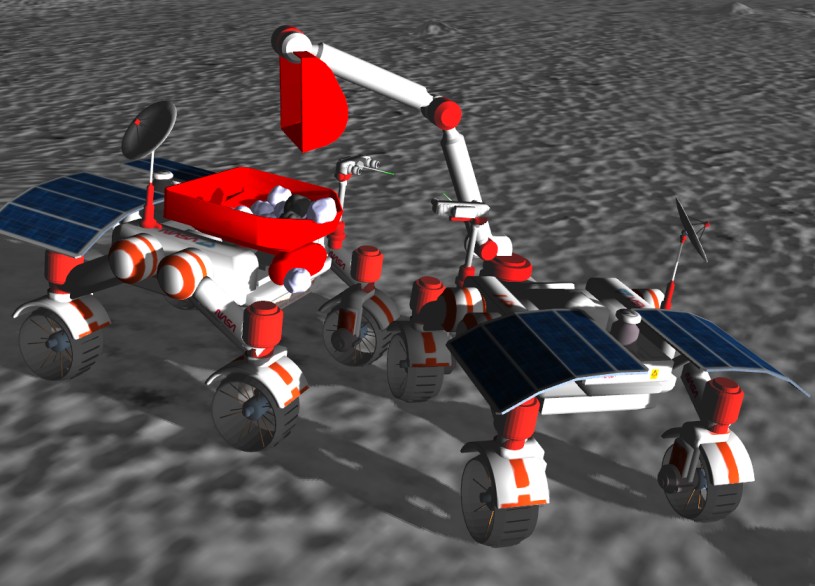}}
    \subfigure{\includegraphics[height=0.35\columnwidth]{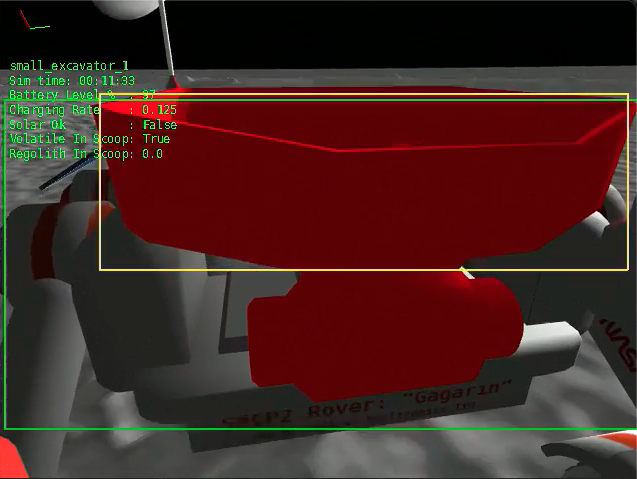}}
    \caption{In the excavation and dumping activity, the object detector on the excavator's camera locates the bin on the hauler to contribute to relative pose estimation. See~\cite{yt_excavation} for a video of the process.}
    \label{fig:excavation}
\end{figure}

With the hauler parked close to the excavator, the digging process begins; illustrated by Fig.~\ref{fig:excavation}. 
The excavator uses vision to estimate its relative location and orientation, which is used to set the scoop angle for resource depositing. 
This process begins with the excavator panning its camera towards the hauler until the bin is identified using YOLOv5, and then LiDAR is used to measure the closest point between hauler's bin and the excavator. All LiDAR measurements are projected into a 2D plane and we only consider object detections that intersect within the bounding box of the hauler's bin.
If the closest point measurement reports that the distance between the two rovers is undesirable, the hauler will readjust its park accordingly. 
The excavator then digs volatiles from ground, and deposits them into the back of the hauler, which continues until the resource patch is depleted.

\section{Robot Coordination}

To facilitate task coordination in multi-robot systems, potential architectures~\cite{archdef} include: a centralised system where all the robots are connected to a central control unit~\cite{matoui2020contribution}, a distributed system where there is no central control and all the robots are equal and autonomous in decision making~\cite{renoux2015decision}, and a decentralised system which is an intermediate between centralised and distributed architectures~\cite{dai2016cooperative}. We opted for a decentralised approach that offers more scalability and higher risk tolerence than a centralised system, whilst being easier to develop and deploy than a distributed system.

Concretely, each rover in a given team is able to autonomously accomplish generic tasks such as localisation, scene understanding, locomotion, as well as specialised ones such as exploration, volatile detection, digging, dumping, parking, etc. 
However, transitioning from one task to the other is done via a centralised coordinator service to facilitate task synchronisation across multiple rovers (e.g. excavator shouldn't dig and deposit resources until the hauler has finished parking). 
Owing to the decentralised nature of our system, should one of the teams break down, the other can continue functioning without any repercussions.


\section{Results}

Qualitative results of the vision-based modules have been provided above. Here, we present some quantitative results.

\paragraph{Localisation}

\begin{figure}[ht]\centering
    \includegraphics[width=1\linewidth]{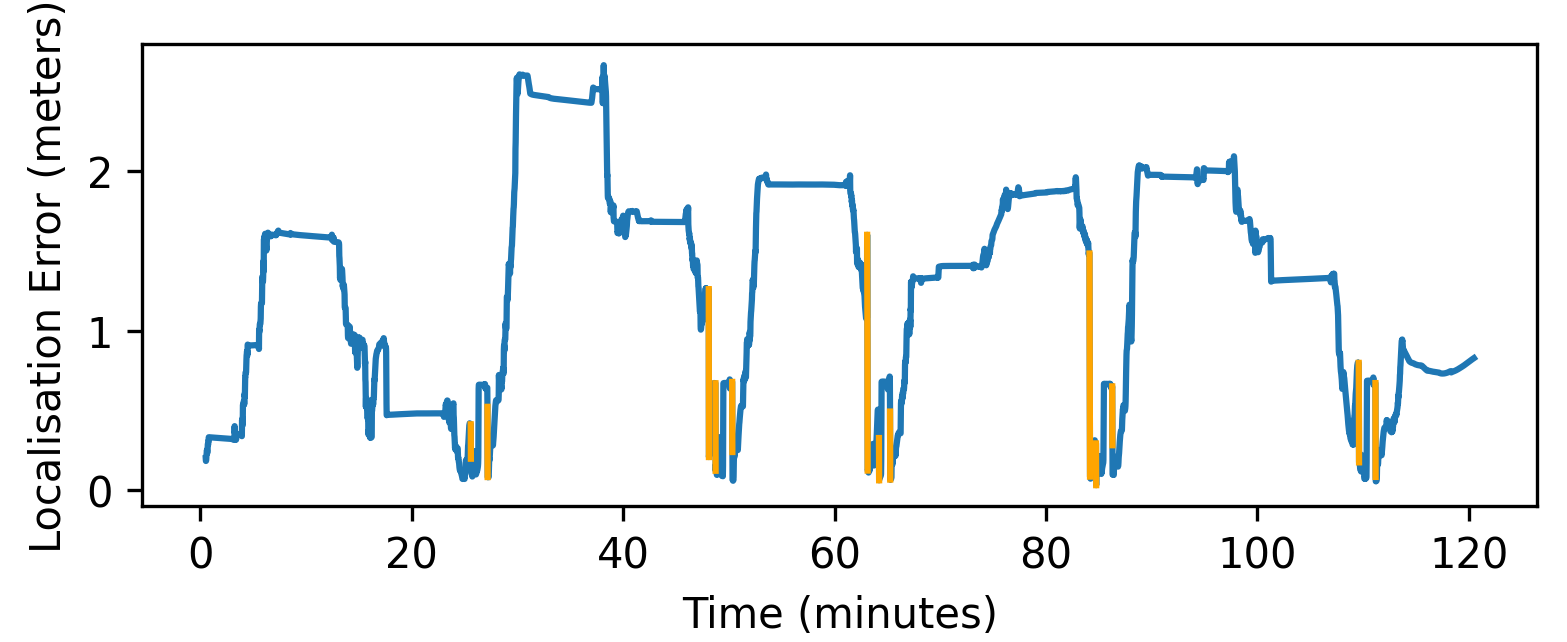}
    \caption{Localisation error of a hauler during a 2-hour simulation run. Localisation resets via visual pose estimation occur frequently throughout the run (highlighted in orange).}
    \label{fig:ekf_error}
\end{figure}

Fig.~\ref{fig:ekf_error} displays the localisation error of a hauler over a 2 hour simulation run, where the error is computed as the difference between the EKF estimate and ground truth position. As shown in the plot localisation error accumulates as the rover moves across the lunar environment. During this run, the error for the hauler reached a maximum of 2.67m. Regular PnP resets ensure that the error stays within acceptable bounds. After resetting, the localisation error was reduced to around 0.2m or less in most cases.

\paragraph{Navigation}

\begin{figure}[ht]\centering
    \includegraphics[width=0.9\linewidth]{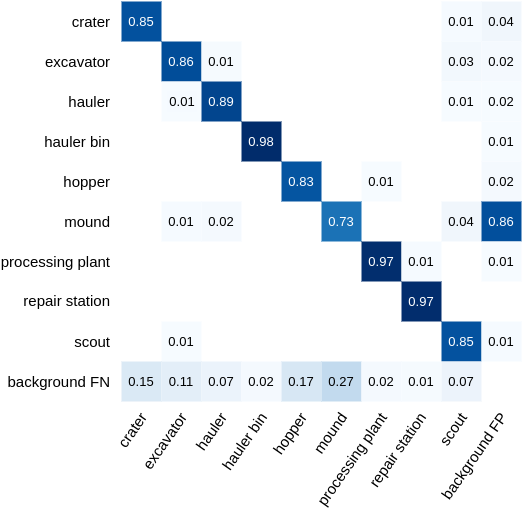}
    \caption{Confusion matrix of our YOLOv5 object detector on testing set.}
    \label{fig:confusion_matrix}
\end{figure}

Semantic scene understanding is a major part of our navigation system. Fig.~\ref{fig:confusion_matrix} shows the confusion matrix of our YOLOv5 object detector model trained on our dataset. The mean testing accuracy of the model is 88\%. Notably, the model generalised successfully to detect small distant mounds not in the training or testing set.

Obstacle detections are added to the persistent local map of the rover at 5 FPS, which was sufficient for all the rovers to achieve fine-grained motor control. Throughout 44 hours of simulation testing, the rovers travelled an accumulated distance of approximately 120km. During these runs, no serious navigational failures due to collisions occurred.

\paragraph{Rover interactions}

As mentioned in Sec.~\ref{sec:interactions}, visual object detection plays a significant role in rover rendezvous and excavation. The same detector model for navigation (i.e., with quantitative results in Fig.~\ref{fig:confusion_matrix}) was used for rover interactions. A more direct measure of success of rover interactions is the amount of volatiles that was extracted by the excavator and deposited into the hauler. Across all resource extraction events \emph{attempted} in the 44 hours of simulation testing, 84.8\% of volatiles were successfully transferred to the hopper. Resource losses were due to rendezvous or deposit inaccuracies, which were more common in challenging terrain (many hills or obstacles in the location of the resource). In overly challenging cases, resource extraction is simply not attempted for the sake of safety; this occurred in about 20\% of the resources discovered by the scout.

\paragraph{Overall results}

\begin{figure}[ht]\centering
    \includegraphics[width=1\linewidth]{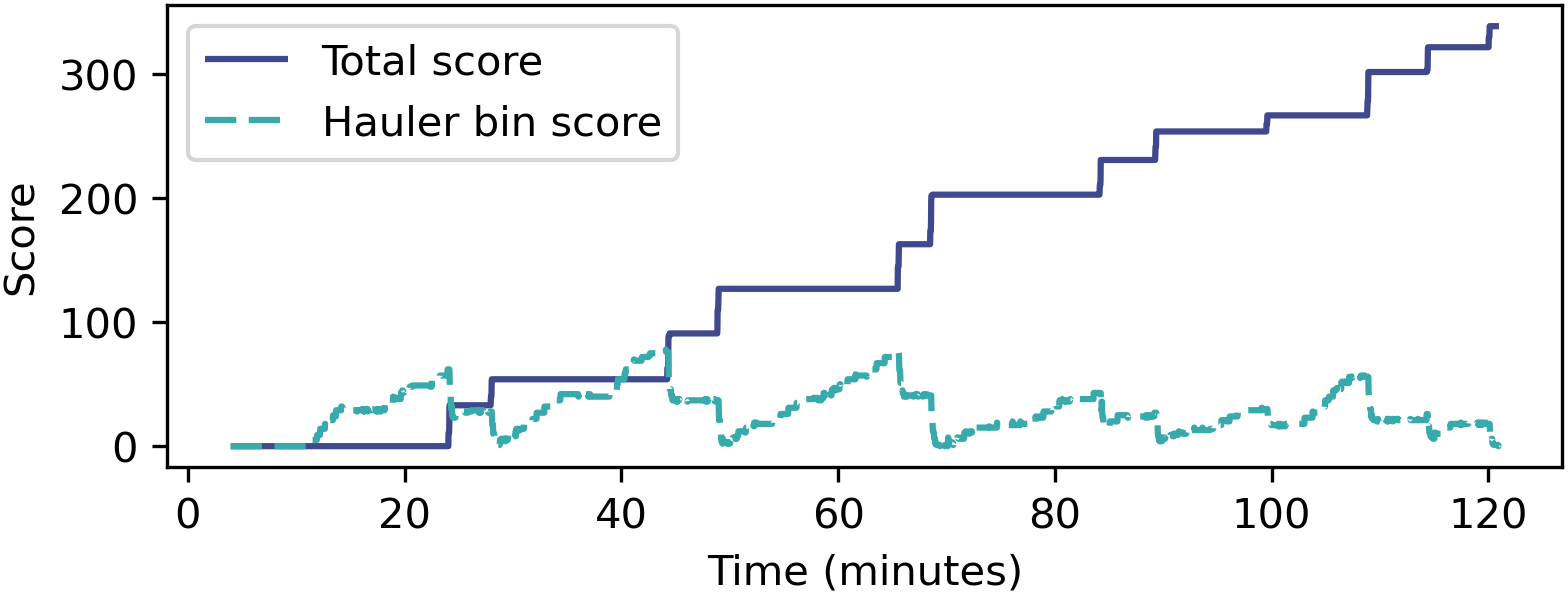}
    \caption{Total score shows the accumulated score throughout a 2 hour run, and hauler bin score shows the number of volatiles present in the bins of the two haulers.}
    \label{fig:score}
\end{figure}

22 simulation runs were performed to evaluate the overall performance of the final system. Each run consisted of 2 hours of simulation time under the competition configuration (see Sec.~\ref{sec:srcp2}). The average, minimum, and maximum number of volatiles extracted during these runs were 266, 163, and 339 respectively. The scores accumulated during a specific 2-hour run are plotted in Fig.~\ref{fig:score}. See also qualitative result in the form of a video recording in~\cite{yt_overall}.

\section{Conclusions}

Our system represents a robust implementation of autonomous space mining in the context of the NASA SRCP2. Guided by robotic vision, our rovers are able to reliably navigate and extract resources from the simulated lunar environment for extended periods. The vision system periodically alleviates localisation drift, as well as to build a persistent map providing semantic scene understanding for use in obstacle avoidance and rover interaction. Interesting future research in the context of robotic vision is to perform VSLAM under the guidance of semantic scene understanding to help alleviate issues due to texture poor terrain, so as to build a semantically meaningful map of the lunar environment.



\section*{Acknowledgements}

We gratefully acknowledge funding from Andy Thomas Centre for Space Resources. We thank the following team members who contributed in various ways towards our solution: John Culton, Hans C. Culton, Alvaro Parra Bustos, Rijul Ramkumar, Shivam Savani, Amirsalar Aryakia, Sam Bahrami, Aditya Pujara and Matthew Michael. James Bockman acknowledges support by the Australian Government Research Program (RTP) Scholarship in conjunction with the Lockheed Martin Australia supplementary scholarship.

\pagebreak

\bibliographystyle{IEEEtran}
\bibliography{refs}

\end{document}